\documentclass[letterpaper, 10 pt, conference]{ieeeconf-iot}  

\IEEEoverridecommandlockouts                              

\usepackage{graphicx} 
\usepackage{amsmath} 
\usepackage{amssymb}  
\usepackage[hidelinks]{hyperref}
\usepackage{subcaption}
\usepackage{comment}
\usepackage[english, ruled, linesnumbered, vlined]{algorithm2e}
\usepackage{booktabs}
\usepackage{bm}
\usepackage{fancyhdr}
\usepackage{url}

\def\iotlogo{\includegraphics[width=2.0cm]{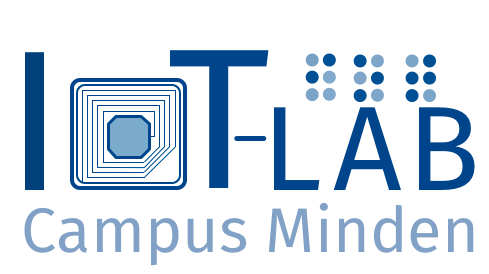}}
\fancypagestyle{iotpagestyle}{
    \fancyhf{}
    
    \fancyfoot[L]{{\rule{\linewidth}{0.4pt}}\\
    \parbox[b][1cm][t]{2cm}{\iotlogo}
    \parbox[b][1cm][t]{15cm}{\footnotesize \bigskip This work originates from the IoT-Lab of Bielefeld University of Applied Sciences, Campus Minden: \url{http://iot-minden.de}}
    }    
}
\title{\LARGE \bf
A Framework for Interactive Teaching of Virtual Borders to Mobile Robots
}

\author{Dennis Sprute$^{1, 2}$,  Robin Rasch$^{1}$, Klaus T{\"o}nnies$^{2}$ and Matthias K{\"o}nig$^{1}$
\thanks{$^{1}$The authors are with Campus Minden, University of Applied Sciences Bielefeld, 32427 Minden, Germany
        {\tt\small {forename.surname}@fh-bielefeld.de}}%
\thanks{$^{2}$The authors are with the Faculty of Computer Science, Otto-von-Guericke University Magdeburg, 39106 Magdeburg, Germany  {\tt\small dennis.sprute@ovgu.de}, 
        {\tt\small klaus@isg.cs.uni-magdeburg.de}}%
}

\begin{document}

\maketitle
\pagestyle{empty}

\begin{abstract}
The increasing number of robots in home environments leads to an emerging coexistence between humans and robots. Robots undertake common tasks and support the residents in their everyday life. People appreciate the presence of robots in their environment as long as they keep the control over them. One important aspect is the control of a robot's workspace. Therefore, we introduce virtual borders to precisely and flexibly define the workspace of mobile robots. First, we propose a novel framework that allows a person to interactively restrict a mobile robot's workspace. To show the validity of this framework, a concrete implementation based on visual markers is implemented. Afterwards, the mobile robot is capable of performing its tasks while respecting the new virtual borders. The approach is accurate, flexible and less time consuming than explicit robot programming. Hence, even non-experts are able to teach virtual borders to their robots which is especially interesting in domains like vacuuming or service robots in home environments.
\end{abstract}

\section{INTRODUCTION}
There is a trend towards service robots in home environments. Robots support the residents of the building by performing tasks, e.g. vacuum cleaning \cite{Hess:2014} or unloading items from the dishwasher \cite{Saxena:2008}. There is a wide range of different tasks that service robots are capable of. Xu et al.~\cite{Xu:2014} identify other functionalities like picking and placing objects, opening doors, meal preparation and cooperative object carrying with humans. Furthermore, they are used in the health-care sector, e.g. as social robots to support elderly people \cite{Gross:2015}. The result is a shared environment between humans and robots.\par

\begin{figure}
	\centering
	\includegraphics[width=0.45\textwidth]{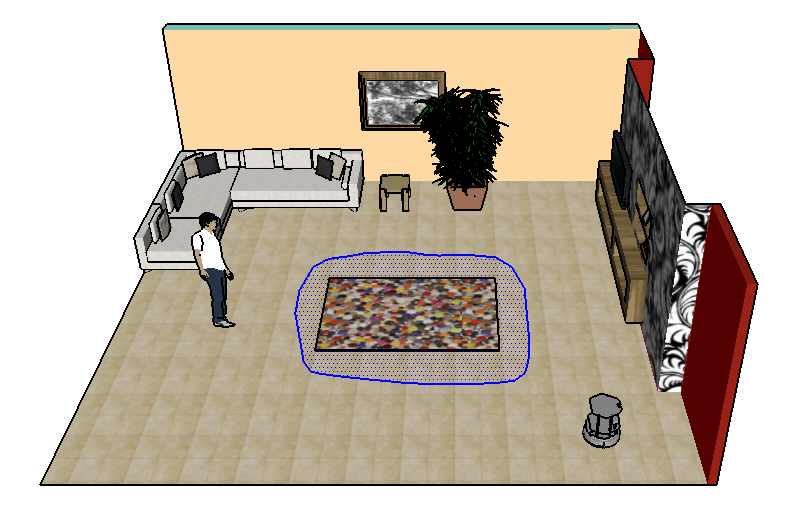}   
	\caption{Motivation scenario to teach virtual borders. A resident wants the robot to exclude the carpet area from its workspace. The blue polygon visualizes the area to be respected by the robot. Hence, the non-expert user has to interact with the robot in order to update its navigation map. }
	\label{fig:motivation}
\end{figure}

The workspace of such a mobile robot is limited by the walls of the building or furniture. Along these physical workspace limitations, it is desirable to further restrict the spatial workspace of a mobile robot, e.g. if there are areas in the building that should not be intruded by a robot due to privacy concerns. Besides, the precise definition of a space for working, such as vacuuming or mopping, plays an important role. An illustration of the problem is shown in Fig. \ref{fig:motivation}. Since there is an expensive carpet on the ground, the resident wants the robot to circumvent it while performing its vacuuming task. In such a case, the robot has to accept \textit{virtual borders} that have to be programmed by the user. Virtual borders are further referred to non-physical borders in the workspace of a robot that have to be respected by the robot while performing its tasks.

The ``classical'' way would be to program the virtual borders into the program flow of the robot, but this is expensive and not feasible for non-expert users. Therefore, it is important that a method for teaching virtual borders is applicable by non-experts without programming skills. Teaching should take little effort to guarantee the acceptance of the users. Additionally, such a method should allow flexible virtual borders teaching, i.e. the user can define any area in the environment according to his needs. Finally, the learned virtual borders need to be as accurate as possible in order to precisely navigate around virtually occupied areas in future navigation tasks. By restricting the workspace of a mobile robot, the user keeps the control over the robot and is able to flexibly define certain areas for working.\par

In this paper, we address the above mentioned problem of \textit{how to teach virtual borders to mobile robots}. Our contribution is twofold: first, we propose a framework for teaching virtual borders to mobile robots working in the two-dimensional plane. It is an interactive approach that can be easily adapted to different human-robot interfaces, such as human gestures. Second, to demonstrate the validity of this framework, we use the framework to implement a teaching method based on visual markers as human-robot interface. We chose visual markers because it is the easiest way to evaluate the proposed framework concerning correctness, accuracy and effort. Our interactive approach differs from other approaches in that it is accurate, flexible and applicable by non-experts.\par      

The remainder of this paper is organized as follows. Next, we give an overview of related works concerning the topic, and we motivate the development of a new framework to teach virtual borders. Subsequently, we formally define the problem and detail the proposed framework. We use this framework to implement a marker-based teaching method and highlight details in the following section. Afterwards, we evaluate the marker-based approach with respect to its correctness, accuracy and teaching effort. Finally, we summarize our work and show future research opportunities.

\section{RELATED WORK}
Mapping and SLAM (Simultaneous Localization and Mapping) are own subareas of robotics that overlap with the topic of this paper. SLAM algorithms try to build maps of the robot's physical environment while simultaneously localizing the robot inside the map~\cite{Thrun:2005}. A history and formal definition of the SLAM problem are given by Durrant-Whyte and Bailey~\cite{Durrant-Whyte:2006}. We refer the interested reader to the article by Cadena et al.~\cite{Cadena:2016} who give a good overview of the evolution of SLAM from the past to the future.\par

In contrast to mapping physical environments, virtual borders have to be mapped in a different way. Sakamoto et al.~\cite{Sakamoto:2009} propose a stroke-based interface on a tablet PC to control the workspace of a vacuum cleaning robot. This technique needs several top-down view cameras in the robot's environment to cover the whole working area. Commercial solutions to the problem of defining the robot's workspace comprise magnetic strips that are placed on the ground to indicate the borders \cite{Neato:2017} or beacon devices to generate infrared signals~\cite{Chiu:2011}. These conic beams will not be crossed by the robot and can be used to block doors or corridors. The disadvantages are intrusiveness, low flexibility and energy consumption. Other explicit methods to incorporate additional information into maps encompass situated dialogues~\cite{Zender:2008} or activity recognition based on wearables~\cite{Li:2012}. These works integrate semantics into maps, but they do not allow the flexible definition of certain areas, e.g. arbitrary polygons inside a room. The flexibility of our approach distinguishes our work from previous works.\par

In contrast to these direct approaches to incorporate additional information into maps, several works focused on the implicit integration of further information, e.g. social information~\cite{Kruse:2013}. O'Callaghan et al.~\cite{O'Callaghan:2011} learn human motion patterns and update the robot's navigational map according to humans' trajectories. In addition, Alempijevic et al.~\cite{Alempijevic:2013} use a map learned from robots' sensors and human trajectory observations to navigate to any goal in the environment. Human motion maps are proposed by Ogawa et al.~\cite{Ogawa:2014} to represent the motion distribution in particular areas. Furthermore, Wang et al.~\cite{Wang:2016} learn about human behaviors associated with areas in the environment. Another approach based on implicit observations is proposed by Papadakis et al.~\cite{Papadakis:2016} who learn the locations of doors and staircases by observing humans' interactions around them. Virtual border teaching, as we use this term throughout the paper, is a certain form of semantic mapping~\cite{Kostavelis:2015} in that we give semantics, \textit{free} and \textit{occupied} space, to certain areas.\par

Although these implicit approaches are more user-friendly compared to explicit techniques, we argue that it is inevitable to use an explicit technique in the case of teaching arbitrary virtual borders. From our experiences, we know that users want robots to avoid certain areas in the environment, and this can only be defined in an explicit way using human-robot interaction. Therefore, we set a focus on the interactive teaching of virtual borders.\par

To allow non-experts the teaching of virtual borders, a user-friendly approach has to be chosen. Along classical robot programming, LfD (Learning from Demonstration) is a technique that deals with teaching robots new skills by (human) demonstrations. Argall et al.~\cite{Argall:2009} present a comprehensive survey of robot LfD and explain the foundations of the technique. The authors categorize different approaches according to the mapping between the teacher and the learner. Teleoperation is used to learn the grasping of objects~\cite{Sweeney:2007} or to perform tasks demonstrated by kinesthetic teaching~\cite{Somani:2013}. If the learner uses its own sensors, but the teacher does not directly control the learner's platform, it is referred to as shadowing. Nehmzow et al.~\cite{Nehmzow:2007} program a robot's movement by demonstration through system identification, and autonomous navigation in complex unstructured terrain is learned in the work of Silver et al.~\cite{Silver:2010}. Other recent applications range from learning handwriting by demonstration \cite{Hood:2015} to controlling a robot arm manipulator \cite{Cai:2016}. The last two applications also use visual markers in their teaching process.

\section{VIRTUAL BORDER TEACHING}
As stated in the previous section, SLAM is used to build maps of physical environments, and there are explcit approaches to restrict the workspace of a mobile robot. However, these approaches are intrusive \cite{Neato:2017}, power-consuming \cite{Chiu:2011} and inflexible in terms of the definition of an arbitrary area~\cite{Zender:2008}, \cite{Li:2012}. Furthermore, implicit methods are not flexible enough to teach arbitrary virtual borders because the approaches do not directly take the users' intentions into account. In order to address these restriction and overcome the shortcomings, we propose a new framework for restricting the workspace of mobile robots in an interactive way. It is inspired by LfD shadowing technique and features correct, accurate, flexible and user-friendly border teaching. By means of user-friendly, we understand an approach that allows teaching by non-expert users with no programming skills.

\subsection{Problem Definition}
In this subsection, we formally define the problem and introduce the notation that is used throughout the paper. The environment of a robot is modeled as an occupancy grid map $M$ consisting of $m \times n$ cells. Each cell contains a probability for the occupancy of the corresponding area in the environment. $M(x, y)$ allows the access of the cell's value at position $x$ and $y$.\par

A two-dimensional mobile robot's pose is defined as a triple $(x, y, \theta)$ comprising the robot's position $(x, y)$ and orientation $\theta$ with respect to a map coordinate frame. While moving through the environment, the mobile robot's pose history up to a certain time $k$ is referred to $X_{0:k}$. The transformation between two consecutive poses $x_{k-1}$ and $x_{k}$ is described by the control vector $u_k$. These notations are consistent with the SLAM formulation of Durrant-Whyte and Bailey~\cite{Durrant-Whyte:2006}.\par

We define a virtual border as a polygon \mbox{$P:=(P_1, P_2, ..., P_n)$} consisting of $n$ points where $P_i \in \mathbb{R}^2, 1\leq i \leq n$. It is the goal to construct a map $M_{posterior}$ given a prior map $M_{prior}$ and a virtual border map $M_{virtual}$. The prior map $M_{prior}$ is constructed from a common mapping algorithm or is a result of a previous teaching process, whereas the virtual border map $M_{virtual}$ is constructed in the interactive teaching process. It is 
defined by a border polygon and a keep off area. The resulting posterior map $M_{posterior}$ will contain physical as well as virtual borders. It can be used as basis for a costmap in future navigation tasks.

\subsection{Framework}
\label{sec:concept}
\begin{figure}
	\centering
	\includegraphics[width=0.45\textwidth]{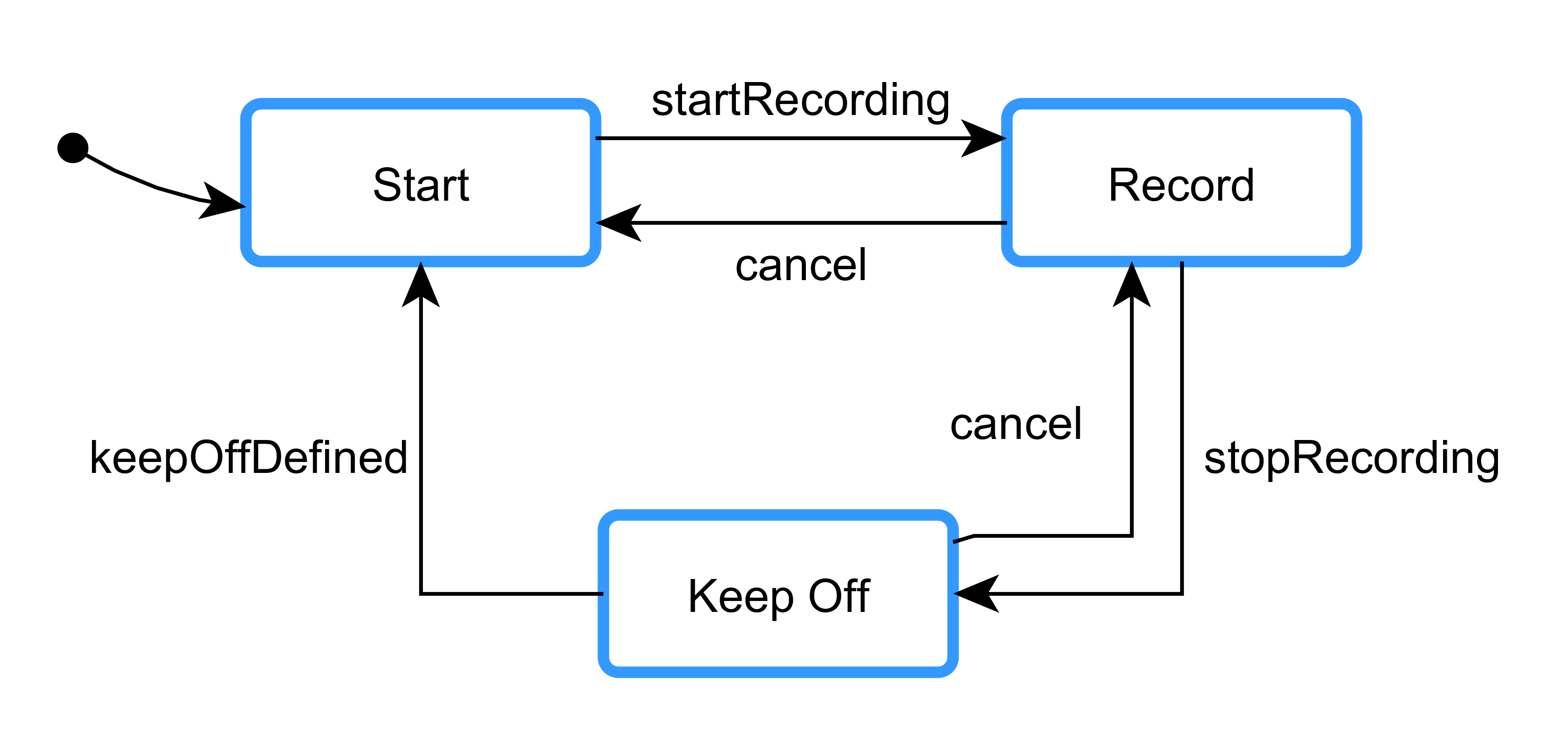}   
	\caption{State diagram of the proposed framework with its three states and transitions.}
	\label{fig:statechart}
\end{figure}

To address the problem of teaching virtual borders in an interactive way, we propose a learning framework based on shadowing where the mobile robot is encouraged to follow a position in space and to record its pose history. The position data are further used to define the virtual border polygon. It is a novel event-based framework that consist of three states representing the states of the teaching process. In order to use this framework, a concrete implementation only has to define the events in an appropriate way, while the implementation of the states can remain the same between different applications. This framework makes it easy to adapt to other human-robot interfaces, such as human gestures or remote control. The transitions between the states are based on events, that are triggered by user interactions, and are visualized in Fig.~\ref{fig:statechart}. The three states are described below:
\begin{enumerate}
	\item \textit{Start} The mobile robot follows a position on the ground plane that is indicated by the human teacher in the teaching process. Note that this is independent of the concrete human-robot interface, e.g. a teacher could use human gestures or a mediator device to provide a position to follow.
	\item \textit{Record} The mobile robot follows the position on the ground plane and records its pose history $X_{a:b}$. The robot enters the state at time $a$ and leaves the state at time $b$.	
	\item \textit{Keep Off} The mobile robot stops recording its pose history $X_{a:b}$, and a virtual border polygon $B$ is extracted from the robot's pose history. Furthermore, the keep off area is defined. There are two possibilities: the inner area of the border polygon $B$ is set to be \textit{occupied}. This keeps the robot away from the inner of the polygon and is useful to exclude an area from the robot's workspace. The other possibility is to declare the inner area of the polygon as \textit{free} and the rest as \textit{occupied}. This is useful to define a working area for the robot. The keep off area is defined by the last known position provided by the user. Finally, the virtual border and their keep off area are integrated into the prior map resulting in a posterior map consisting of physical and virtual borders.
\end{enumerate}

\section{MARKER-BASED VIRTUAL BORDER TEACHING}
In order to validate the proposed framework, we use the framework to implement a teaching method based on visual markers to guide the mobile robot. We chose visual markers because they are widely used for preliminary experiments, e.g. \cite{Hood:2015}, \cite{Cai:2016},  \cite{Stein:2013}, and provide an easy human-robot interface. Since the framework is easily adaptable, it will serve as basis for more intuitive human-robot interfaces in the future.

\subsection{Method}
\begin{figure*}
		\centering
        \begin{subfigure}[b]{0.3\textwidth}
                \centering
                \includegraphics[width=\textwidth]{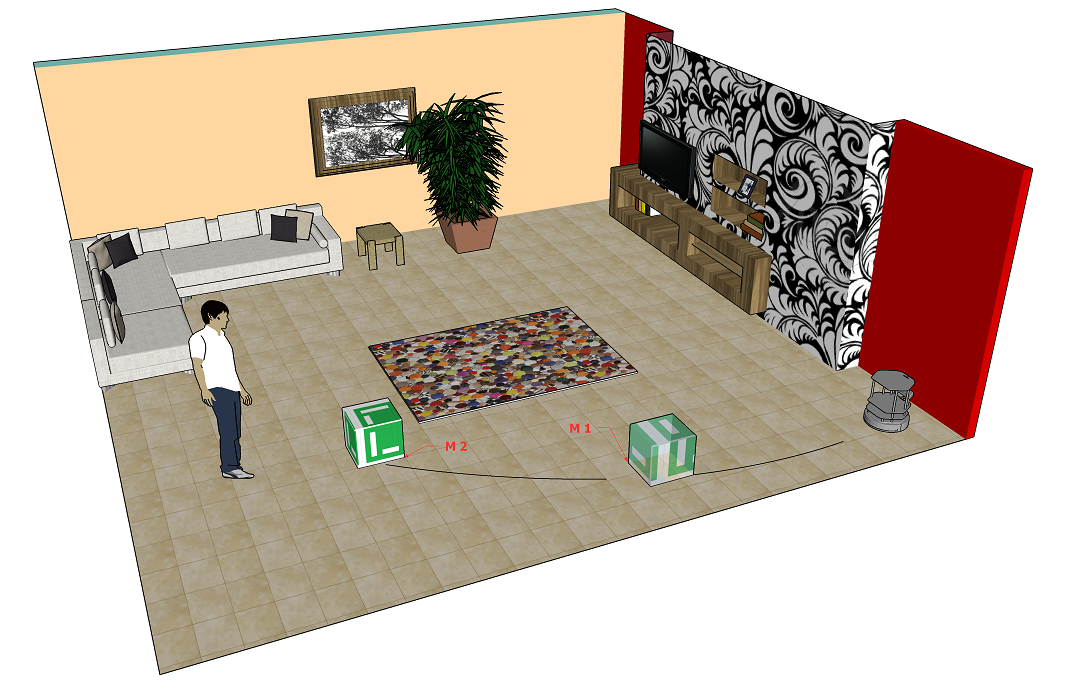}
                \caption{Start}                
        \end{subfigure}        
        \centering
        \begin{subfigure}[b]{0.3\textwidth}
                \centering
                \includegraphics[width=\textwidth]{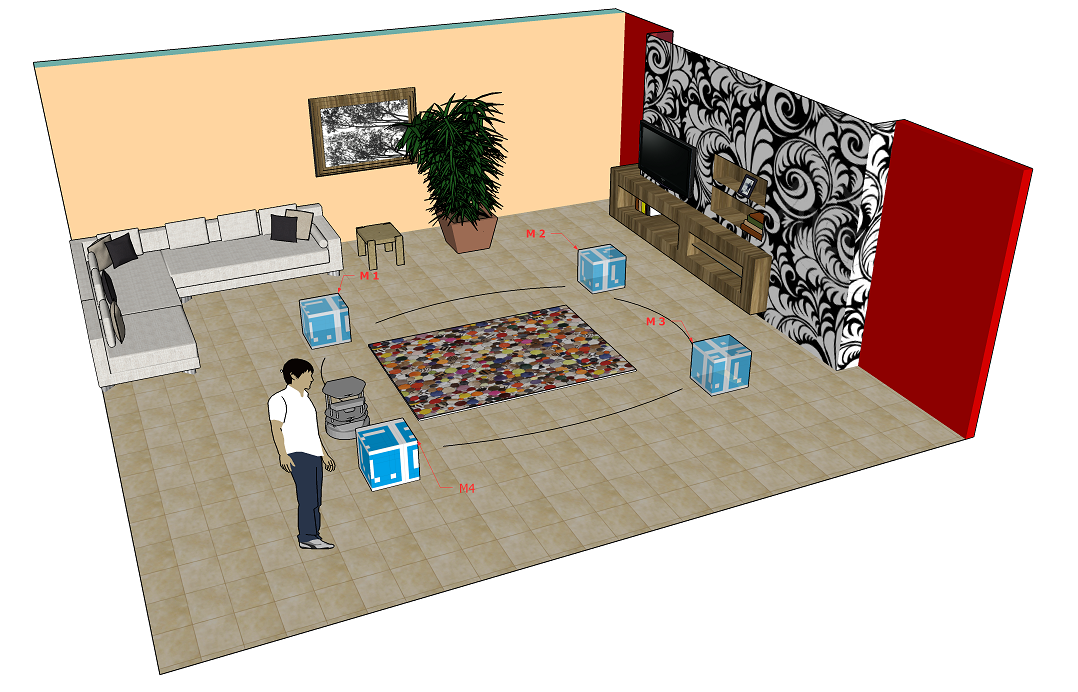}
                \caption{Record}                
        \end{subfigure}          
        \centering
        \begin{subfigure}[b]{0.3\textwidth}
                \centering
                \includegraphics[width=\textwidth]{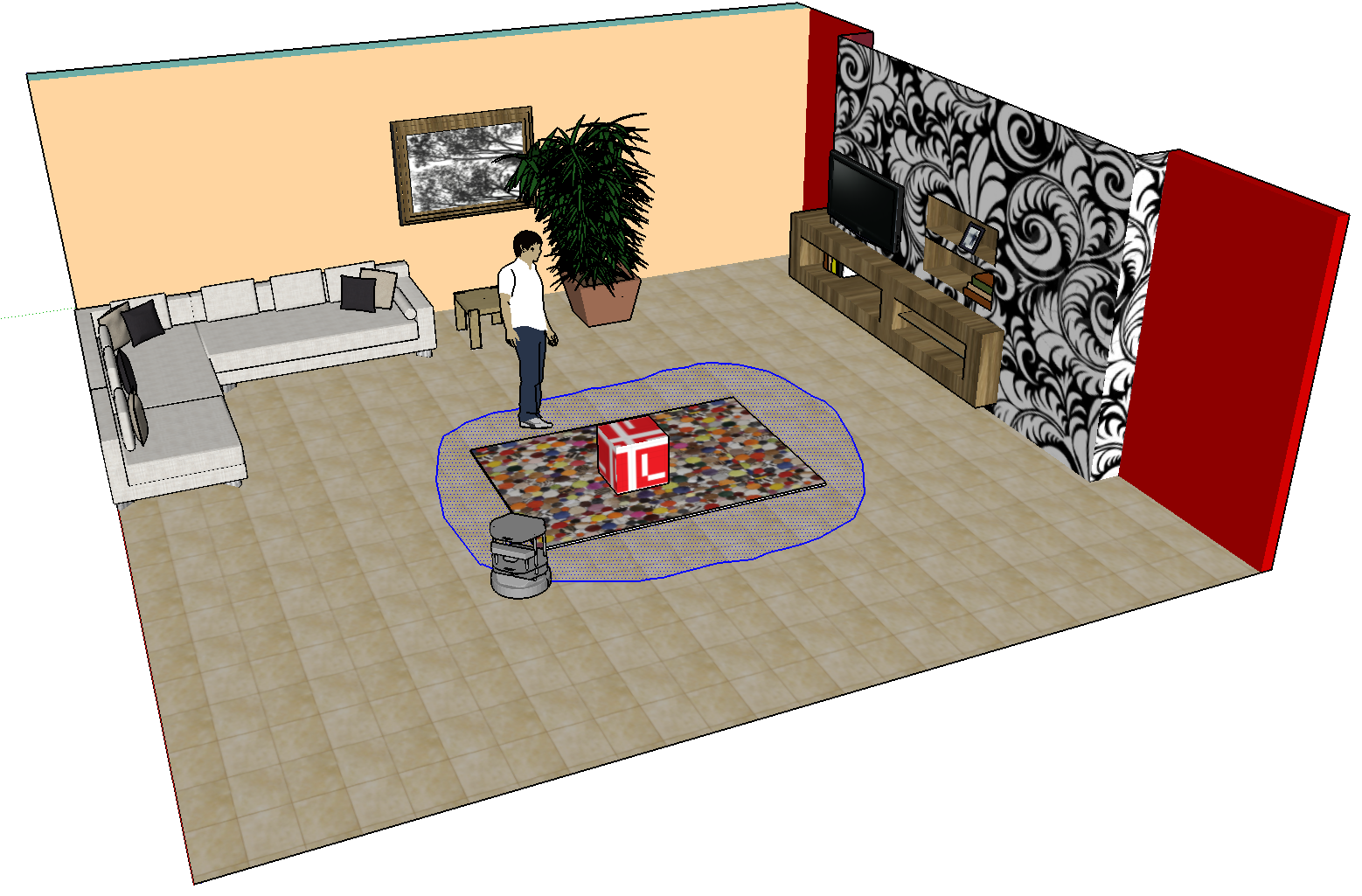}
                \caption{End}                
        \end{subfigure}                   
        \caption{States of the marker-based teaching method. (a) illustrates the guidance of the robot to the start position of the border learning process. (b) shows the teaching of the virtual borders where the robot records its ego-motion data. The final position of the marker and the ego-motion data are used to define the virtual border as shown in (c).} 
        \label{fig:markerLearning}
\end{figure*}
We use three different markers in the teaching process. A change of the marker corresponds to a transition between two states. The complete marker-based teaching method is illustrated in Fig. \ref{fig:markerLearning} where each marker id is represented by a different color. It is inspired by the previous illustration example of defining a virtual border around a carpet. At the beginning of the process, there is an initial map of the environment, and the mobile robot is only restricted in its workspace by the physical borders of the room. To start the teaching process, the user places the \textit{green} marker in the robot camera's field of view, and the robot starts following the marker. The \textit{green} marker is used to guide the robot to a start position. By changing the id of the marker to \textit{blue}, the robot continues following the marker and starts recording its pose history $X_{a:b}$. Recording starts at time $a$ and ends at time $b$ when leaving the state. Subsequently, the \textit{red} marker indicates the end of the border learning procedure, and the robot's position data as part of the robot's pose history $X_{a:b}$ are extracted to define the virtual border. Finally, the keep off area has to be defined. Since the robot does not have to cross the carpet, the final rotation of the robot is adjusted by placing the marker in the inner of the polygon. In the case of defining a working area as opposed to a keep off area, the final robot rotation has to point away from the polygon. The teaching process is successfully finished if the robot does not move for a time $t$ which is currently set to $t=10$ seconds. Additionally, the teaching can be canceled by switching to marker id \textit{green} from any other state.

\subsection{Implementation Details}
The following requirements have to be fulfilled to realize the introduced method of marker-based border teaching:
\begin{enumerate}
	\item Mapping: A prior map representation of the environment has to be available. It should be stored as an occupancy grid map with probabilities ranging from $[0, 1]$. Unknown areas are marked with a $-1$. In the current implementation, mapping is performed using a laser scanner and ROS gmapping package which is based on a particle filter to solve the SLAM problem~\cite{Grisetti:2007}. This map is further referred to the prior map $M_{prior}$ of the environment. Note that $M_{prior}$ can also be a resulting map from previous teaching processes.
	\item Localization: The robot needs to be localized with respect to the map's coordinate frame. In the current implementation, adaptive Monte Carlo localization~\cite{Fox:2003} is chosen which is implemented in the ROS amcl package.
	\item Sensors: The robot needs a monochrome camera to acquire gray-scale input images. It has to be mounted on the robot pointing towards the robot's $x$-axis (forward direction). The camera's intrinsic parameters need to be known to obtain the 3D-position of the marker with respect to the camera frame.
\end{enumerate}

Virtual border teaching is performed by following visual markers. For this purpose, a 10 cm $\times$ 10 cm $\times$ 10 cm marker cube with one ArUco marker \cite{Garrido-Jurado:2014} per site is created. Three different ids are generated and used by the teaching person. Every marker has a fixed size, provides two-dimensional image features and is represented by its four corner points. The ArUco library performs marker detection and identification to obtain markers from an input image. If intrinsic camera parameters are given additionally, the pose of the marker with respect to the camera can be estimated. Therefore, depth information of the marker's position can be obtained with a single camera. These marker positions are used to guide the robot towards the marker. Distance information of the marker acts as stop condition and controls the speed of the robot.\par

While following the \textit{blue} marker, the robot records its pose. Motion data $u_k$ are acquired by wheel odometry, and the new pose of the robot $x_k$ is updated by applying $u_k$ to the previous pose of the robot $x_{k-1}$. Since the robot is localized inside the map, the poses are transformed from the robot's odometry coordinate frame into the map's coordinate frame.\par

\begin{figure}
	\centering
	\includegraphics[width=0.45\textwidth]{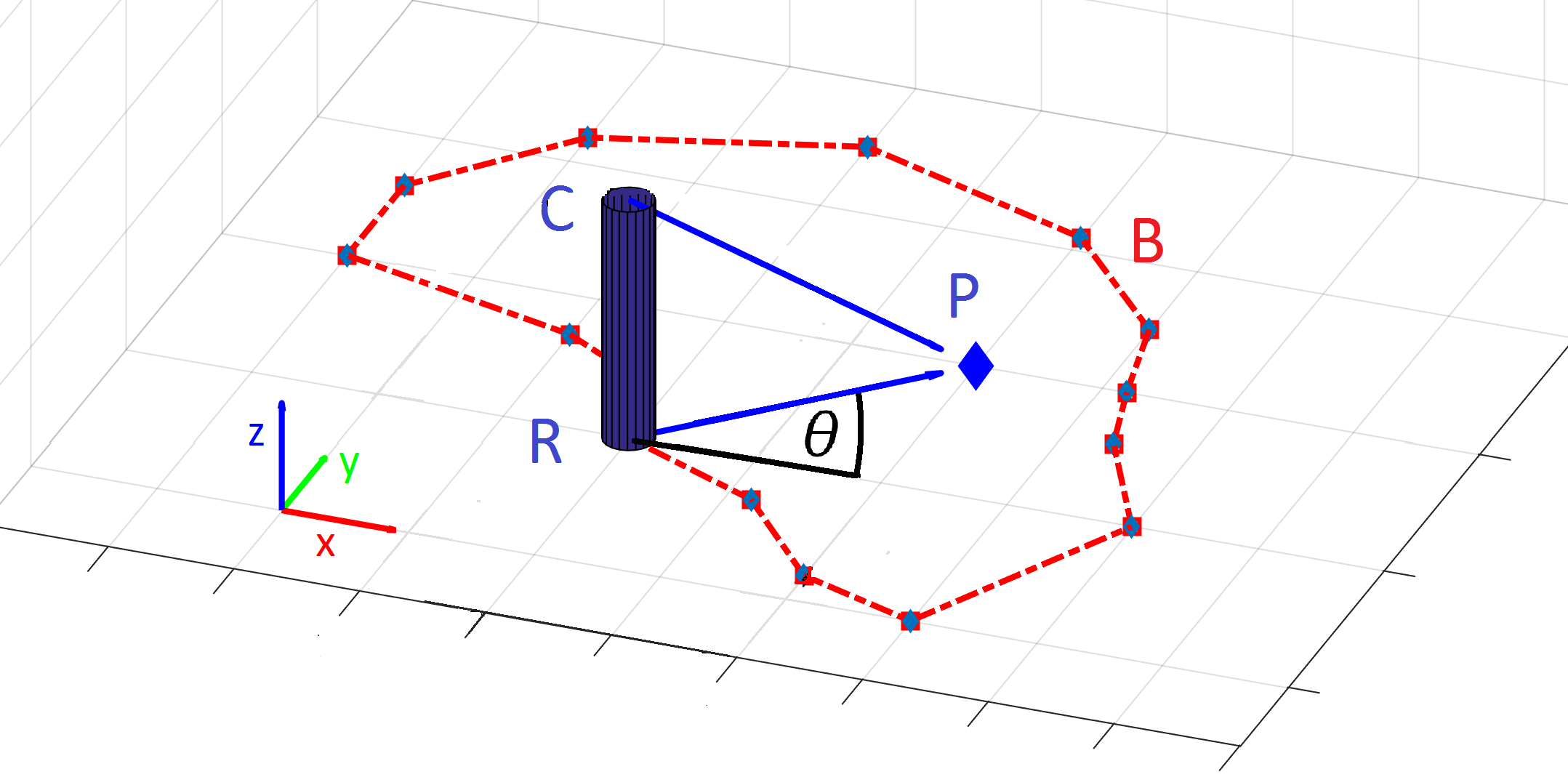}   
	\caption{Illustration of the underlying geometry for determining the keep off area. The current robot's position $R$ and its camera position $C$ are depicted as a blue cylinder. $B$ denotes the virtual border polygon and $P$ the last known marker position.}
	\label{fig:keepOffArea}
\end{figure}

When the marker id changes from \textit{blue} to \textit{red}, the user has the possibility to rotate the robot around its $z$-axis to indicate the keep off area. If the robot points towards the inner of the polygon, the inner area is considered as \textit{occupied}, otherwise as \textit{free}. The underling geometry is illustrated Fig. \ref{fig:keepOffArea}.

The decision is based on the last marker's position $P$. Since the robot turns around its $z$-axis pointing towards the marker, the vector $\overrightarrow{RP}$ from the robot's current position to the marker is described by the robot's orientation $\theta$. This vector is defined as follows:
\begin{equation}
\begin{split}
d = \sqrt{\parallel\overrightarrow{CP}\parallel^2-\parallel\overrightarrow{RC}\parallel^2}\\
\overrightarrow{RP} =  d \cdot (cos(\theta), sin(\theta))^T
\end{split}
\end{equation}
The distance between the camera and the marker $\parallel\overrightarrow{CP}\parallel$ is determined by utilizing the intrinsic camera parameters and the known size of the marker. Subsequently, the position of the marker $P$ is calculated:
\begin{equation}
P = R + \overrightarrow{RP}
\end{equation}
This point $P = (P_x, P_y)$ is used in a standard point-in-polygon test $\Phi(P_x, P_y, B) \in \{true, false\}$  with the virtual border polygon $B$ to define the keep off area. The test returns true if the point $(P_x, P_y)$ lies inside the polygon $B$. \par

Finally, the map containing the virtual borders $M_{virtual}$ can be defined as follows:
\begin{equation}
M_{virtual}(x, y)=
\begin{cases}
	1& if\ \Phi(x, y, B) \wedge \Phi(P_x, P_y, B)\\
	0& if\ \neg\Phi(x, y, B) \wedge \Phi(P_x, P_y, B)\\
	1& if\ \neg\Phi(x, y, B) \wedge \neg\Phi(P_x, P_y, B)\\	
	0& if\ \Phi(x, y, B) \wedge \neg\Phi(P_x, P_y, B)\\
\end{cases}
\end{equation}
After creating a virtual border map $M_{virtual}$, we integrate it into the prior map $M_{prior}$ of the environment resulting in the posterior map $M_{posterior}$: 
\begin{equation}
M_{posterior}(x, y)= \bm{max} (M_{prior}(x, y), M_{virtual}(x, y))
\end{equation}
This is necessary to restrict the workspace of the mobile robot in future navigation tasks by serving as basis for a 2D costmap.\par

The whole system is implemented as a ROS package~\cite{Quigley:2009} and can be deployed on any mobile robot with a forward mounted monochrome camera to perceive the marker cube. Additionally, the above mentioned requirements concerning mapping and localization have to be met.

\section{EVALUATION}
We evaluated the marker-based teaching method concerning three criteria that serve as measurements for the previously mentioned system requirements. First, the correctness of the method is evaluated showing the overall functionality of the system. Subsequently, the accuracies of the learned posterior maps are assessed by comparing them with a self-recorded dataset containing maps with integrated virtual borders. These maps have been recorded previously and are further referred to as ground truth data. Finally, the time to teach a border is measured, which serves as an indicator for the effort of the teaching method. The dataset contains ten different maps of a 6.1 m $\times$ 3.5 m lab environment with virtual borders' lengths ranging from 4 m to 13 m. The virtual borders are defined by different polygons that are convex and non-convex. The teaching process is performed five times for each map resulting in 50 trials. The robot starts from different positions in each trial to show the general applicability of the approach. Each trial is performed by guiding the robot along a predefined virtual border. We used a TurtleBot V2 with a  front-mounted camera and laser scanner to obtain the experimental results. Before performing the experiments, a physical map of the lab environment is created using ROS gmapping package and the robot is localized using ROS amcl package. All experiments are performed on maps with a resolution of 2.5 cm per pixel.

\subsection{Correctness}
\begin{figure*}
\centering
        \begin{subfigure}[b]{0.49\textwidth}
                \centering
                \includegraphics[width=\textwidth]{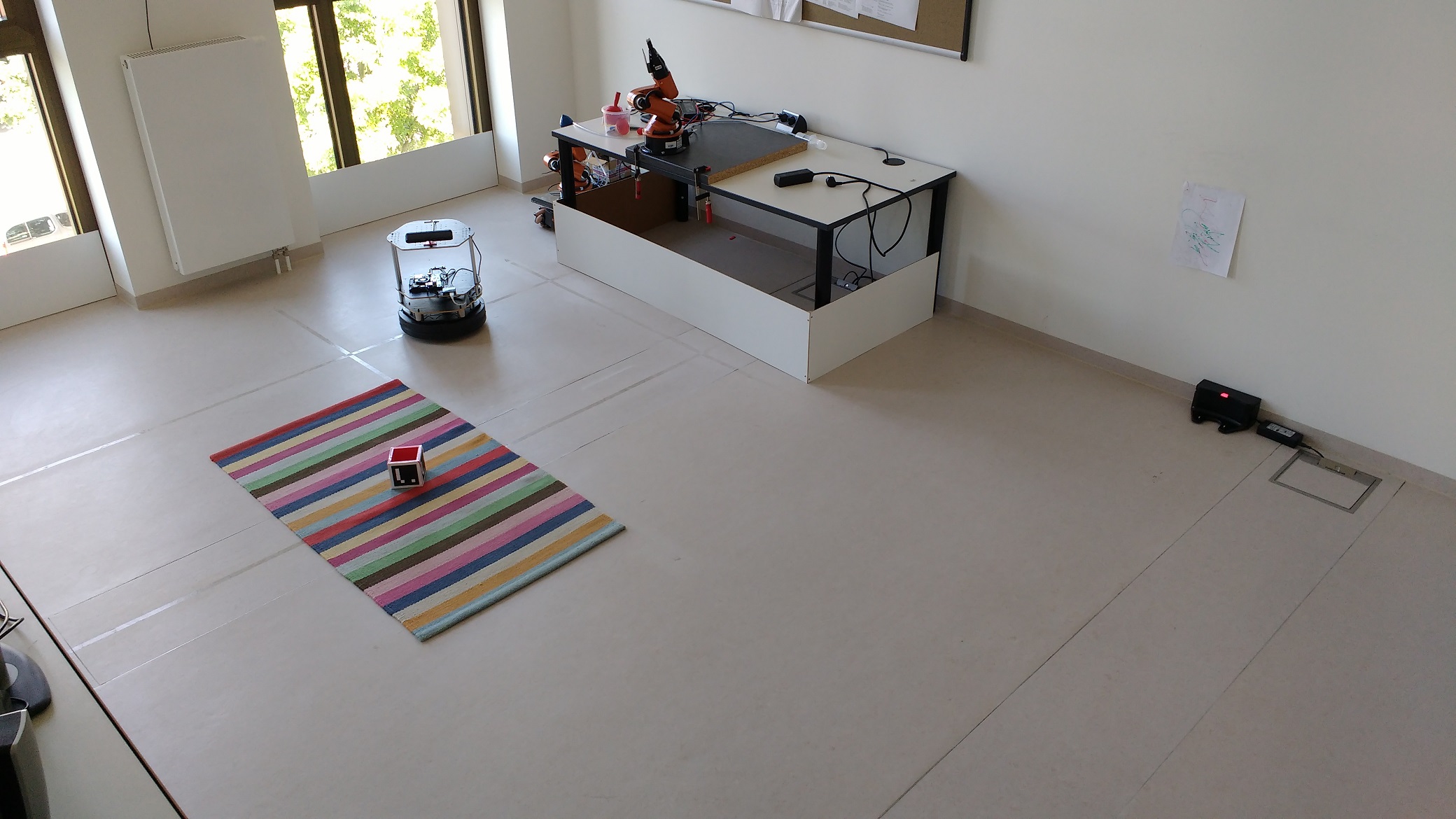}
                \caption{Lab environment}                
        \end{subfigure} 
		\centering
        \begin{subfigure}[b]{0.45\textwidth}
                \centering
                \includegraphics[width=\textwidth]{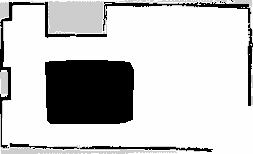}
                \caption{Occupancy grid map}                
        \end{subfigure}                             
        \caption{Marker-based teaching example. (a) shows the operating environment of the robot with a carpet placed on the ground. This area should not be crossed by the robot while performing its task, e.g vacuuming. (b) depicts the occupancy grid map after the teaching process. The cells corresponding to the area of the carpet are marked as \textit{occupied}, while the rest of the map remains the same as the prior map.} 
        \label{fig:mapAndLab}
\end{figure*}

This section is intended to provide results concerning the correctness of the proposed method. Therefore, we take up again the previously mentioned scenario of teaching a virtual border around a carpet. It is the goal to declare the area of the carpet as occupied, whereas the rest of the map should stay the same. A user takes the marker cube and guides the robot around the carpet according to Fig. \ref{fig:markerLearning}. The teaching process finishes successfully by placing the marker cube on the carpet to indicate the keep off area. The results of this teaching process are visualized in Fig. \ref{fig:mapAndLab}. The left figure shows an image of the lab environment, and the right image depicts the resulting map containing the physical as well as the virtual border around the carpet area. This area is marked in black and indicates an occupied area. It can be used as input map for further navigation tasks, and the robot will not enter the area of the carpet. The results demonstrate the general functionality of the proposed framework as specified before.

\subsection{Accuracy}
The evaluation of the accuracy aims to provide a measure for the accuracy of the learned virtual borders compared to ground truth data. A high accuracy value entails precisely learned borders. While performing the experiments, the generated posterior maps are stored and associated with their ground truth maps. To assess the accuracy, the Jaccard index between two corresponding maps $A$ and $B$ is calculated:
\begin{equation}
J(A, B) = \dfrac{|A \cap B|}{|A \cup B|}
\end{equation}
$|A \cap B|$ is the number of cells that are equal in $A$ and $B$, while $|A \cup B|$ is defined by the total amount of cells in a map. The index is in the range of $[0, 1]$ with a high value corresponding to a high accuracy of the maps. Table \ref{tab:accuracy} shows the Jaccard indices for the different maps:
\begin{table}[htbp]
  \centering
  \caption{Accuracy results of the experiments}
    \begin{tabular}{rrrrr}
    \toprule
          &       & \multicolumn{3}{c}{Jaccard index} \\    
    Map   & Length [in m] & Minimum   & Maximum   & Average \\\midrule
    Map 1 & 4     & 99.0\% & 99.6\% & 99.2\% \\
    Map 2 & 5     & 98.0\% & 99.1\% & 98.5\% \\
    Map 3 & 6     & 97.0\% & 98.2\% & 97.6\% \\
    Map 4 & 7     & 97.9\% & 98.7\% & 98.3\% \\
    Map 5 & 8     & 97.4\% & 98.9\% & 98.1\% \\
    Map 6 & 9     & 97.7\% & 98.2\% & 98.0\% \\
    Map 7 & 10    & 97.2\% & 98.3\% & 97.8\% \\
    Map 8 & 11    & 96.9\% & 98.7\% & 97.9\% \\
    Map 9 & 12    & 96.8\% & 97.6\% & 97.1\% \\
    Map 10 & 13    & 96.0\% & 97.7\% & 96.9\% \\\bottomrule
    Average & 8.5   & 97.4\% & 98.5\% & 97.9\% \\  
    \end{tabular}%
  \label{tab:accuracy}%
\end{table}%

The average accuracy in the experiments is 97.9\% which underlines the high accuracy of the teaching method. There is a small range between the minimum and the maximum showing the stability of the system. Small errors occur due to the human-robot interaction and localization inaccuracies. Additionally, the accuracy decreases minimally with an increasing border length. The longer the length of the virtual border is, the higher is its affect on the accuracy measure. Since there are no comparable systems with an accuracy evaluation, these results can serve as a baseline for future approaches.

\subsection{Effort}
\begin{figure}
	\centering
	\includegraphics[width=0.48\textwidth]{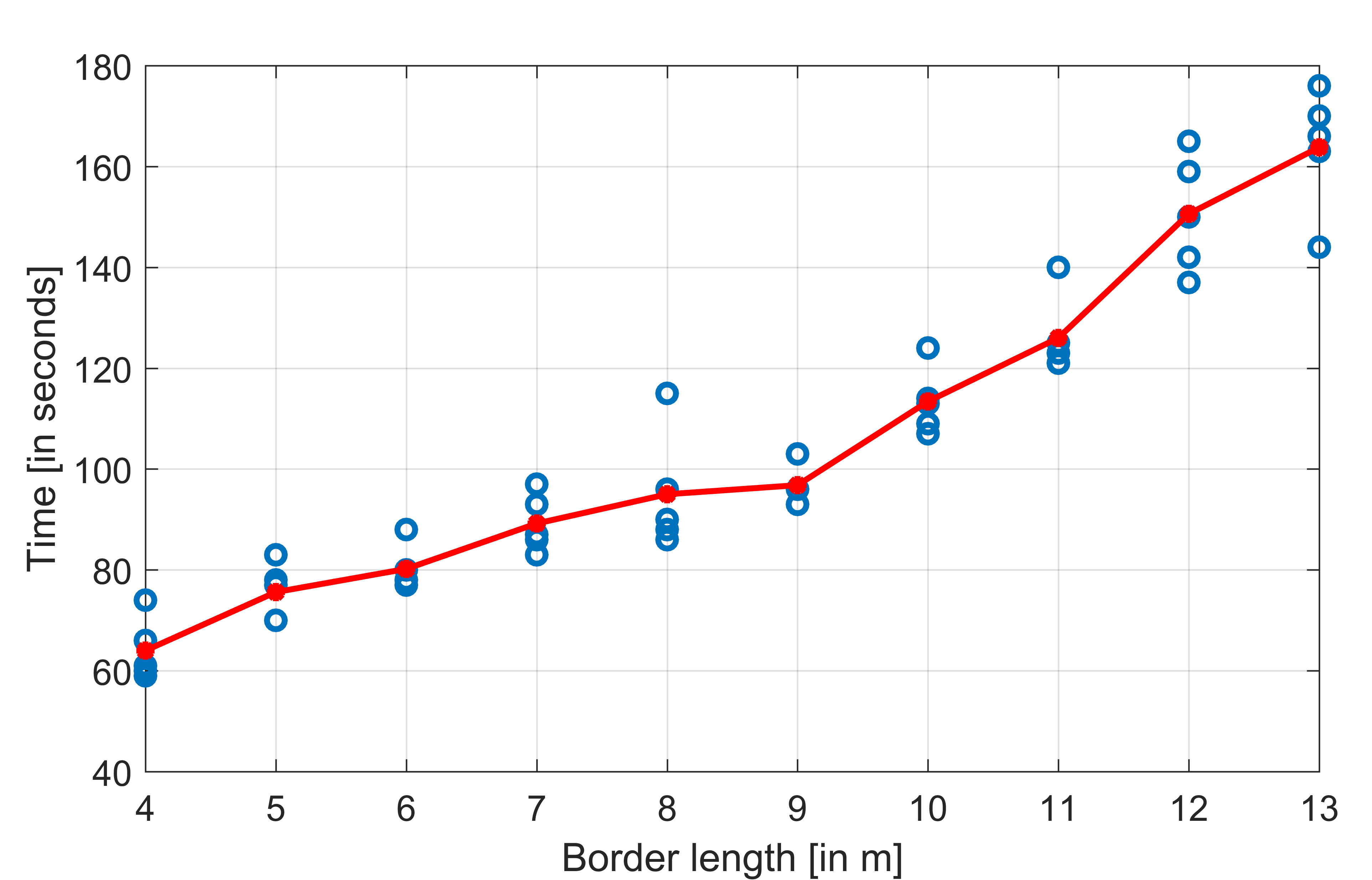}   
	\caption{Teaching time dependent on the border length. There are blue circle for each trial of the experiment and five circles per map with borders ranging from 4 to 13 meters. The average teaching time per map is depicted with a red circle, and the trend of the averages depending on the border length is visualized by a red line.}
	\label{fig:timeEval}
\end{figure}

To evaluate the effort of the teaching method, we consider the time to teach virtual borders. It is the time the robot needs to generate a new map containing the virtual borders. We exclude the time of moving to the start position (state \textit{Start} in Fig. \ref{fig:statechart}) from the time measurement because it does not give information about the teaching process. Fig. \ref{fig:timeEval} gives an overview of the teaching time dependent on the length of the virtual border.

There is a linear relationship between the teaching time and the length of a border which is caused by the nature of the proposed framework. A mobile robot follows a marker, and the longer it follows the marker, the more time it takes. The minimal teaching time for a 4 meters long border is 59 seconds, and 176 seconds are the maximum for teaching a 13 meters long border. Since there is only a small deviation from the average teaching time, the time to teach virtual borders can be easily estimated.

\subsection{Discussion}
The experimental results in this section demonstrate the correctness of the proposed teaching framework and the high accuracies of the learned posterior maps. Furthermore, it takes little effort to teach virtual borders and the framework is applicable by non-expert users. The effort of teaching virtual borders gives an indicator for the usability of the system, but this also strongly depends on the chosen human-robot interface. We chose markers as basis for interaction with the robot to validate the proposed framework with respect to the correctness, accuracy and teaching effort. In order to deploy the framework in a real-world scenario, we currently implement more intuitive human-robot interfaces in combination with the framework. 

\addtolength{\textheight}{-1.0cm}   

\section{CONCLUSIONS \& FUTURE WORK}
We proposed a new framework for teaching virtual borders to mobile robots in an interactive way. The framework is inspired by shadowing technique and addresses non-experts as users. Furthermore, it is easily adaptable to different human-robot interfaces, such as human gestures. We evaluated the validity of the proposed framework by implementing a method based on the robot's on-board camera and visual markers as interaction interface. The learned virtual borders will be respected by the robot in future navigation tasks. Experimental results revealed the correctness of the system, a high accuracy as well as a good teaching time. Furthermore, the user is able to flexibly define polygonal areas as virtual borders. Maps containing arbitrary virtual borders can be defined by successively iterating the teaching process. Such a framework is especially interesting for residents in home environments to restrict their mobile robots workspaces in an interactive way. Therefore, this framework serves as a good basis for future developments in this area.\par

Thanks to the easily adaptable framework, future work will focus on implementations using intuitive human-robot interfaces to increase the usability of the system. Currently, we work on the realization using a laser pointer as mediator device and human gestures to guide the robot. We are particular interested in the user's perspective in the teaching process. Therefore, we plan to conduct a comprehensive study to investigate the influence of different human-robot interaction methods on the teaching process. Furthermore, we will focus on the implementation of different border types in order to define the workspace more flexible. Currently, only polygons are supported but other types such as curves or lines separating a room will be explored. Finally, the development of an adequate  feedback system showing the learned virtual borders is also planned.

\section*{ACKNOWLEDGMENT}
This work is financially supported by the German Federal Ministry of Education and Research (BMBF, Funding number: 03FH006PX5).

\bibliography{bibo}   
\bibliographystyle{IEEEtran}

\end{document}